\documentclass[runningheads]{llncs}
\usepackage[dvipsnames]{xcolor}
\usepackage{graphicx}
\usepackage{subfigure}

\usepackage{tikz}
\usepackage{comment}
\usepackage{amsmath,amssymb} 
\usepackage{color}
\usepackage{pgfplots}
\pgfplotsset{compat=1.17}

\usepackage[accsupp]{axessibility}  

\usepackage{lipsum}

\usepackage{multirow}
\usepackage{bbding}
\usepackage{float}
\usepackage{hyperref}

\begin{document}
\pagestyle{headings}
\mainmatter
\def\ECCVSubNumber {3723}  

\title{MTTrans: Cross-Domain Object Detection with Mean Teacher Transformer}

\titlerunning{MTTrans: Cross-Domain Object Detection with Mean Teacher Transformer}
%

\author{Jinze Yu$^1$\and
Jiaming Liu$^2$\and
Xiaobao Wei$^2$\and
Haoyi Zhou$^1$\and
Yohei Nakata$^3$\and
\\Denis Gudovskiy$^3$\and
Tomoyuki Okuno$^3$\and
Jianxin Li$^1$\and
\\Kurt Keutzer$^4$\and
Shanghang Zhang$^{2}$\thanks{Corresponding Author}}

\authorrunning{J. Yu et al.}
%
\institute{$^1$Beihang University \quad $^2$Peking University\\
$^3$Panasonic Holdings Corporation
\quad $^4$University of California, Berkeley \\
\email{yujinze@buaa.edu.cn, shanghang@pku.edu.cn}}
\maketitle

\begin{abstract}

Recently, DEtection TRansformer (DETR), an end-to-end object detection pipeline, has achieved promising performance. However, it requires large-scale labeled data and suffers from domain shift, especially when no labeled data is available in the target domain. 
To solve this problem, we propose an end-to-end cross-domain detection Transformer based on the mean teacher framework, \textbf{MTTrans}, which can fully exploit unlabeled target domain data in object detection training and transfer knowledge between domains via pseudo labels. 
We further propose the comprehensive multi-level feature alignment to improve the pseudo labels generated by the mean teacher framework taking advantage of the cross-scale self-attention mechanism in Deformable DETR. 
Image and object features are aligned at the local, global, and instance levels with domain query-based feature alignment (DQFA), bi-level graph-based prototype alignment (BGPA), and token-wise image feature alignment (TIFA).
On the other hand, the unlabeled target domain data pseudo-labeled and available for the object detection training by the mean teacher framework can lead to better feature extraction and alignment.
Thus, the mean teacher framework and the comprehensive multi-level feature alignment can be optimized iteratively and mutually based on the architecture of Transformers.
Extensive experiments demonstrate that our proposed method achieves state-of-the-art performance in three domain adaptation scenarios, especially the result of Sim10k to Cityscapes scenario is remarkably improved from 52.6 mAP to 57.9 mAP. 
Code will be released \href{https://github.com/Lafite-Yu/MTTrans-OpenSource}{here}.

\end{abstract}

\section{Introduction}

Object detection is one of the fundamental computer vision tasks which has been improved dramatically in the last decades. Methods based on Convolutional Neural Networks (CNN)~\cite{ren2015faster,redmon2016you,liu2016ssd} achieve satisfying results but rely heavily on certain hand-crafted operations and are not fully end-to-end. 
Recently, Transformer-based approaches~\cite{carion2020end,zhu2020deformable} have been introduced as a promising one-stage detector. 
Although Transformer-based detectors have shown superior detection and generalization performance compared with CNN-based ones~\cite{bai2021transformers,zhang2022delving}, they still suffer significant performance degradation caused by domain shift or variation of data distribution when tested in scenarios with domain gaps. 
These motivate the study of unsupervised domain adaptive (UDA) on transformer object detectors~\cite{wang2021exploring,zhang2021dadetr}.

The pivot of UDA is to deal with domain shifts between source and target domains. It empowers the learned model to be transferred from the source to the unlabeled target domain. 
Most previous works are based on cross-domain feature alignment techniques, which rarely use target data in object detection training due to the lack of labels, thus causing insufficient use of data. 
In this paper, inspired by the prevalent mean teacher mechanism and the usage of pseudo labels in semi-supervised learning (SSL) tasks~\cite{sohn2020simple,xu2021end} and other domains~\cite{chen-etal-2021-pseudo}, we innovatively design MTTrans, a cross-domain detection Transformer based on the mean teacher framework.
By extending the mean teacher framework to UDA, we can sufficiently exploit unlabeled target data and transfer the knowledge between domains via pseudo labels.

However, directly incorporating the mean teacher framework to UDA object detection often results in degressive performance: the framework is proposed initially to solve the SSL task, in which the labeled and unlabeled data are in the same data distribution. 
The goal of UDA is to remove domain-specific components from the extracted features, and the mean teacher does not have such an ability. 
Such domain-specific features increase the divergence between source and target features and degrade the quality of pseudo labels generated by the mean teacher.
Therefore, it is crucial to exploit cross-domain feature alignments to learn domain-invariant features and improve the quality of pseudo labels in the target domain, which is the critical factor for better domain adaptation.

\begin{figure} [t]
    \centering
    \subfigure[Cityscapes to Foggy]{
        \begin{minipage}[t]{0.33\linewidth}
            \begin{tikzpicture}[scale=0.45]
                \centering
            	\begin{axis}[
            	    symbolic x coords={person, rider, car, truck, bus, train, mcycle, bicycle, mean, 0},
            	    xtick={person, rider, car, truck, bus, train, mcycle, bicycle, mean},
            	    x tick label style={rotate=45, anchor=east, align=center},
            	    ylabel=AP50,
            	    axis lines*=left,
            	    ymajorgrids = true,
            	    ybar interval=0.5,
            	    ymin=0,
            	    legend style={at={(0.5,-0.2)},anchor=north,legend columns=-1},
            	]
                	\addplot[draw=ProcessBlue, fill=ProcessBlue] 
                	coordinates{
                		(person,47.7) (rider,49.9) (car,65.2) (truck,25.8) (bus,45.9) (train,33.8) (mcycle,32.6) (bicycle,46.5) (mean,43.4) (0,0)
                	};
                	\addplot[draw=BrickRed, fill=BrickRed] 
                	coordinates{
                		(person,46.5) (rider,48.6) (car,62.6) (truck,25.1) (bus,46.2) (train,29.4) (mcycle,28.3) (bicycle,44.0) (mean,41.3) (0,0)
                	};
                	\addplot[draw=Apricot, fill=Apricot] 
                	coordinates{
                		(person,38.8) (rider,45.9) (car,57.2) (truck,29.9) (bus,50.2) (train,51.9) (mcycle,31.9) (bicycle,40.9) (mean,43.3) (0,0)
                	};
                	\addplot[draw=YellowGreen, fill=YellowGreen] 
                	coordinates{
                		(person,37.3) (rider,39.1) (car,44.2) (truck,17.2) (bus,26.8) (train,5.8) (mcycle,21.6) (bicycle,35.5) (mean,28.5) (0,0)
                	};
                	\addlegendentry{MTTrans(Ours)}
                    \addlegendentry{SFA}
                    \addlegendentry{ViSGA}
                    \addlegendentry{Def DETR}
                \end{axis}
            \end{tikzpicture}
        \end{minipage}%
    }%
    \subfigure[Cityscapes to BDD100k]{
        \begin{minipage}[t]{0.33\linewidth}
            \begin{tikzpicture}[scale=0.45]
                \centering
            	\begin{axis}[
            	    symbolic x coords={person, rider, car, truck, bus, mcycle, bicycle, mean, 0},
            	    xtick={person, rider, car, truck, bus, mcycle, bicycle, mean},
            	    x tick label style={rotate=45, anchor=east, align=center},
            	    ylabel=AP50,
            	    axis lines*=left,
            	    ymajorgrids = true,
            	    ybar interval=0.5,
            	    ymin=0,
            	    legend style={at={(0.5,-0.2)},anchor=north,legend columns=-1},
            	]
                	\addplot[draw=ProcessBlue, fill=ProcessBlue] 
                	coordinates{
                		(person,44.1) (rider,30.1) (car,61.5) (truck,25.1) (bus,26.9) (mcycle,17.7) (bicycle,23.0) (mean,32.6) (0,0)
                	};
                	\addplot[draw=BrickRed, fill=BrickRed] 
                	coordinates{
                		(person,40.2) (rider,27.6) (car,57.5) (truck,19.1) (bus,23.4) (mcycle,15.4) (bicycle,19.2) (mean,28.9) (0,0)
                	};
                	\addplot[draw=YellowGreen, fill=YellowGreen] 
                	coordinates{
                		(person,38.9) (rider,26.7) (car,55.2) (truck,15.7) (bus,19.7) (mcycle,10.8) (bicycle,16.2) (mean,26.2) (0,0)
                	};
                	\addlegendentry{MTTrans(Ours)}
                    \addlegendentry{SFA}
                    \addlegendentry{Def DETR}
                \end{axis}
            \end{tikzpicture}
        \end{minipage}%
    }%
    \subfigure[Sim10k to Cityscapes]{
        \begin{minipage}[t]{0.33\linewidth}
            \begin{tikzpicture}[scale=0.45]
                \centering
            	\begin{axis}[
            	    symbolic x coords={0, car, 1},
            	    xtick={car},
            	    x tick label style={rotate=45, anchor=east, align=center},
            	    ylabel=AP50,
            	    axis lines*=left,
            	    ymajorgrids = true,
            	    ybar,
            	    ymin=0,
            	    legend style={at={(0.5,-0.2)},anchor=north,legend columns=-1},
            	]
                	\addplot[draw=ProcessBlue, fill=ProcessBlue] 
                	coordinates{
                		(0,0) (car,57.9) (1,0)
                	};
                	\addplot[draw=BrickRed, fill=BrickRed] 
                	coordinates{
                		(0,0) (car,52.6) (1,0)
                	};
                	\addplot[draw=Apricot, fill=Apricot] 
                	coordinates{
                		(0,0) (car,49.3) (1,0)
                	};
                	\addplot[draw=YellowGreen, fill=YellowGreen] 
                	coordinates{
                		(0,0) (car,47.4) (1,0)
                	};
                	\addlegendentry{MTTrans(Ours)}
                    \addlegendentry{SFA}
                    \addlegendentry{ViSGA}
                    \addlegendentry{Def DETR}
                \end{axis}
            \end{tikzpicture}
        \end{minipage}
    }%
    \caption{The performance of our proposed MTTrans compared with previous works on different datasets. Our method consistently outperforms the SOTA baseline with a large margin on the scene adaptation and the synthetic to real adaptation, and the performance of each individual category.}
    \label{fig:intro_res}
    \vspace{-10pt}
\end{figure}
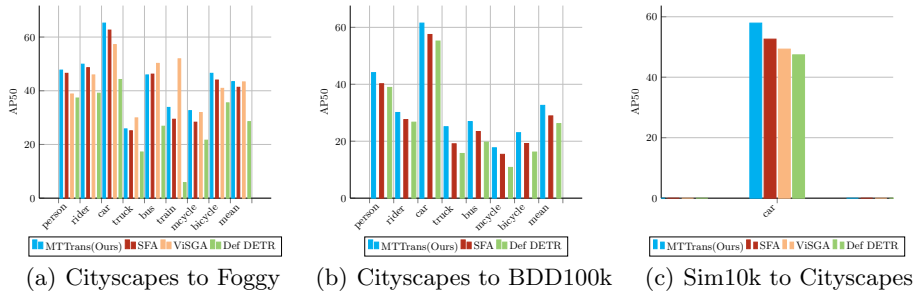

To address the issues above, we propose \textbf{MTTrans}, an end-to-end cross-domain detection Transformer framework, and the first Transformer-based study that utilizes the mean teacher structure to transfer knowledge between domains.
MTTrans transfer domain-invariant task-relevant knowledge between domains and better utilize the unlabeled data via \textbf{pseudo label generation by the mean teacher framework} and the \textbf{comprehensive multi-level cross-domain feature alignment} based on the cross-scale attention mechanism in Deformable DETR~\cite{zhu2020deformable}.
With pseudo label generation, the unlabeled target domain data are pseudo-labeled and available in the object detection training. 
In contrast, in previous works, such data are rarely used due to the lack of labels.
The multi-level feature alignment aligns both images and object proposals features at the global, instance, and local levels to enhance the reliability and the quality of the generated pseudo labels under domain shift.
On the other hand, with target domain images in object detection training, better feature extraction and alignment can be achieved.
Therefore, the mean teacher framework and the feature alignment act as a whole and can be optimized iteratively and mutually. 

The comprehensive feature alignment strategy includes \textbf{domain query-based feature alignment (DQFA)} to align image and object features at the global level, \textbf{bi-level graph-based prototype alignment (BGPA)} which builds prototypes with the object features, and a simple \textbf{token-wise image feature alignment (TIFA)} technique for the local-level image features. 
Inspired by SFA~\cite{wang2021exploring}, 
DQFA extracts a feature for each of the entire image and all the object proposals via cross-scale attention and shrinks global domain gaps in scene layout. However, it still suffers from the domain shift caused by object visual layout changing.
Therefore, we then propose the novel BGPA technique to build and aggregate prototypes for object features based on their visual similarity, which can better learn categorical and spatial correlations (since objects in the same category and spatially connected tend to be visually similar~\cite{rezaeianaran2021seeking}) and achieve more accurate alignment.
Finally, TIFA is performed on the local-level image tokens.
The quality of pseudo labels is largely improved with the comprehensive multi-level feature alignment, as shown in Fig.\ref{fig:pseudo_label_vis}.
A more detailed visualization analysis will be performed in Section~\ref{sec:4.4}.

Finally, the proposed MTTrans framework achieves promising performance on three challenging domain adaptation benchmarks, as shown in Fig.~\ref{fig:intro_res}. 
In the weather adaptation scenario, MTTrans outperforms both end-to-end and two-stage detection algorithms by improving the result to 43.4 mAP (Cityscapes to Foggy Cityscapes). 
In the scene adaptation scenario, the proposed method surpasses the previous state-of-the-art result~\cite{wang2021exploring} by +3.7 mAP, improving the result to 32.6 mAP (Cityscapes to BDD100k). 
In the synthetic to real adaptation scenario, we achieve an improvement of +5.3 mAP (Sim10k to Cityscapes) over the previous state-of-the-art (SFA~\cite{wang2021exploring}), achieving 57.9 mAP.





\section{Related Work}

\subsection{Object Detection}

Object detection is one of the fundamental tasks of computer vision~\cite{he2017mask,kirillov2019panoptic,zheng2015scalable}.
Recently, CNN-based methods with large-scale labeled training data have become the mainstream object detection approaches, categorized into stronger two-stage methods~\cite{ren2015faster,lin2017feature,yang2019reppoints} and faster, lighter one-stage methods~\cite{redmon2016you,liu2016ssd,tian2019fcos}. 
However, these methods extremely depend on handcrafted components, notably the non-maximum suppression (NMS) post-processing, and thus cannot be trained end-to-end. 
It was recently achieved by DETR~\cite{carion2020end} and its follow-up work Deformable DETR~\cite{zhu2020deformable}, with the vision Transformers~\cite{vaswani2017attention}. 
Deformable DETR proposes a novel deformable multi-head attention mechanism to provide sparsity in attention and multi-scale feature aggregation without feature pyramid structure, allowing for faster training and better performance. 
In this work, we choose Deformable DETR as the base detector for its simple yet powerful working flow and the great potential of the attention mechanism for cross-domain feature extraction.

\subsection{Unsupervised domain adaptive object detection}
Domain Adaptive Faster R-CNN~\cite{chen2018domain} is the first work to study domain-adaptive object detection. 
Most later works follow the cross-domain feature alignment with the adversarial training approach. 
These works propose to aggregate image or instance features based on their categorical predictions~\cite{xu2020exploring,xu2020cross} or spatial correlations~\cite{xu2020cross,cai2019exploring}, and the features are aligned hierarchically at one or some levels in global, local, instance and category levels~\cite{xu2020exploring,saito2019strong,xu2020cross,cai2019exploring,luo2021unsupervised}.
More recent works include PICA~\cite{zhong2022pica}, which focuses on Few-shot Domain Adaptation, and Visually Similar Group Alignment (ViSGA)~\cite{rezaeianaran2021seeking}, which proposes to aggregate instance features based on their visual similarity with similarity-based hierarchical agglomerative clustering, which achieves state-of-the-art performance on some benchmarks.
Other methods focus on applying other domain adaptation techniques or other base detectors, such as Mean Teacher with Object Relations (MTOR)~\cite{cai2019exploring} and Unbiased Mean Teacher (UMT)~\cite{deng2021unbiased}. 
Regarding the Transformer-based models, Sequence Feature Alignment (SFA)~\cite{wang2021exploring} proposes a domain adaptive end-to-end object detector based on Deformable DETR. 
DA-DETR~\cite{zhang2021dadetr} proposes an alignment technique based on convolutions and spatial and channel attention for DETR. 
CDTrans~\cite{xu2021cdtrans} and TVT~\cite{yang2021tvt} are proposed for cross-domain classification.
In this paper, we propose cooperating the mean teacher framework with multi-level cross-domain feature alignment based on Deformable DETR to achieve end-to-end UDA object detection.

\section{Method}

This section introduces our MTTrans framework for domain adaptive detection Transformer. In unsupervised domain adaptation (UDA), the data include the labeled source images and the unlabeled target images. Our goal is to transfer object detection task-specific domain-invariant knowledge from the source to the target domain and make the target domain pseudo-labeled and thus available for object detection training with the mean teacher framework and the multi-level source-target feature alignment.

\begin{figure*}[ht]
\begin{center}
	\includegraphics[width=\textwidth]{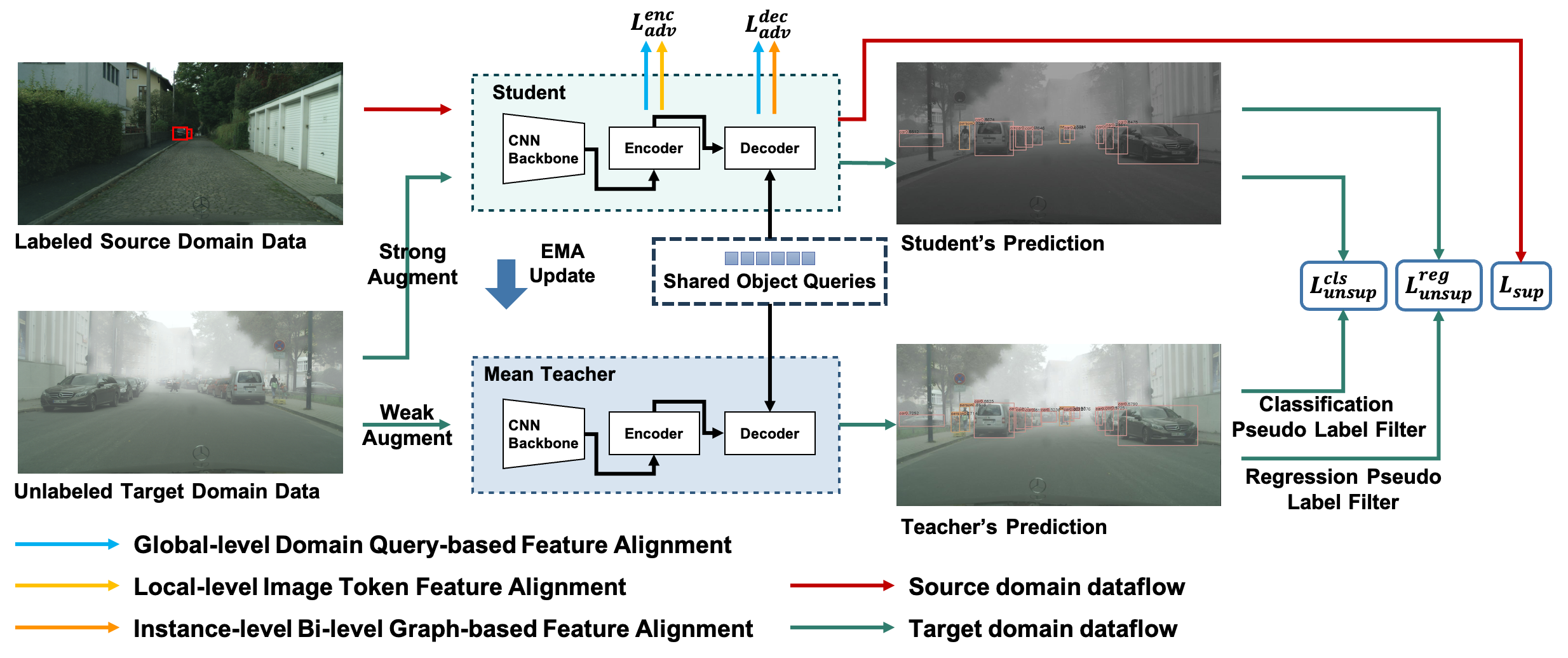}
\end{center}
\caption{The framework of MTTrans, composed of a \textit{student model} which is the actual task model, and a temporal ensembled version of the student model called the \textit{teacher model}. The teacher model is updated by EMA of the student model to generate pseudo labels for the target domain. To improve the quality of the pseudo labels on the target domain and reduce the source-target discrepancy, we additionally design multi-level feature alignment strategies.}
\label{fig:overall_framework}
\end{figure*}

In Section~\ref{sec:3.1}, based on the mean teacher, we propose an end-to-end cross-domain detection Transformer framework, \textbf{MTTrans}, which learns similarities between the source and target domain with pseudo labels. 
We then propose multi-level source-target feature alignment in Section~\ref{sec:3.2} to address the domain shift problem and further improve the reliability of the pseudo labels generated by the mean teacher for the target domain. 
Finally, in Section~\ref{sec:3.3}, we elaborate on the training policy of our framework. 
The overall pipeline is shown in Fig.~\ref{fig:overall_framework}.

\subsection{Mean Teacher-based Knowledge Transfer Framework}
\label{sec:3.1}

MTTrans is based on Deformable DETR~\cite{zhu2020deformable} and is derived from the mean teacher framework, consisting of two models with the same architecture and identical initialization weights. 
The student model is updated with back-propagation, while the teacher model is updated by student's weights with exponential moving average (EMA). 
Thus, the teacher model can be considered as multiple temporal ensembled student models: for weights of the teacher model $\theta'_{t}$ at time step $t$, it is the EMA of successive student's weights $\theta_t$:

\begin{equation}
\label{eq:1}
     \theta'_{t} = \alpha \theta'_{t-1} + (1-\alpha) \theta_{t}
\end{equation}

where $\alpha$ is a smoothing coefficient hyperparameter. All of the teacher's weights are updated according to Eq.~\ref{eq:1}, except for object query embeddings, which are kept the same between the two models to further enhance consistency between them. 
The object queries are trainable embeddings, initialized with the normal distribution at the start of the training procedure. 
Then we use the temporally ensembled teacher model to guide the student model's training in the target domain via pseudo labels.

As a crucial factor for the teacher-student framework, pseudo labels are tactfully generated in our work. 
Object proposals with any foreground category score higher than a pre-defined threshold are assigned as pseudo labels. 
In addition, the unlabeled target images are strongly augmented for the student model (denoted as $I_{tgt}$) and weakly augmented for the teacher model (denoted as $I'_{tgt}$)~\cite{sohn2020fixmatch}. 
In such a way, the teacher model's prediction can be more accurate, enabling the student model to learn from the generated pseudo labels. 
Although directly applying the mean teacher to the UDA task improves the results, the pseudo labels generated on the target domain are still of low quality because of the data distribution shift. The temporal ensembled model tends to accumulate errors and collapse. 

\subsection{Multi-level Cross-Domain Adversarial Feature Alignment}
\label{sec:3.2}

\begin{figure*}[t]
\begin{center}
	\includegraphics[width=\textwidth]{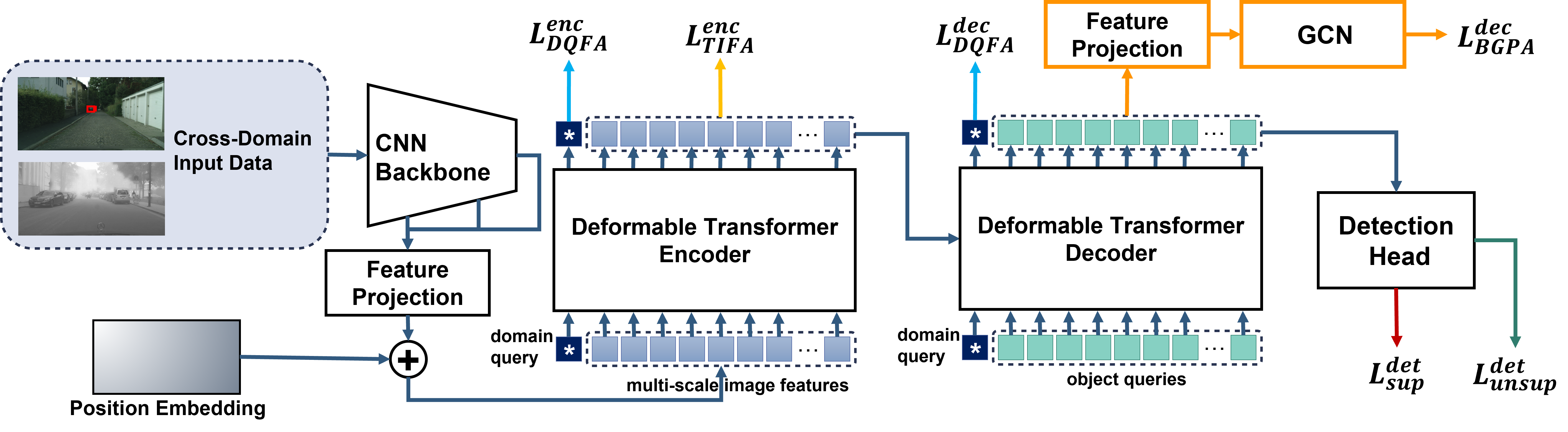}
\end{center}
\caption{The detailed structure of the student model, which is composed of a detection Transformer backbone, and the proposed multi-level feature alignment.}
\label{fig:alignment}
\end{figure*}

To address the issue above, we design the comprehensive multi-level cross-domain feature alignment based on the cross-scale self-attention mechanism of Deformable DETR~\cite{zhu2020deformable} to strengthen the reliability of the pseudo labels generated when faced with domain shift.
The alignment is performed on different parts and at three levels of the student model, including \textbf{domain query-based feature alignment (DQFA)} for global-level image and instance features on encoder and decoder outputs, \textbf{bi-level graph-based prototype alignment (BGPA)} for instance-level object proposal features on decoder outputs, and \textbf{token-wise image feature alignment (TIFA)} for local-level image features on encoder outputs, as shown in Fig.~\ref{fig:alignment}.

\subsubsection{Global-level Domain Query-based Feature Alignment (DQFA).}
Inspired by the global memory~\cite{guo2019star,devlin2018bert} and the sparse attention mechanism~\cite{zhou2021informer,child2019generating}, we adopt DQFA in SFA~\cite{wang2021exploring} to extract and align global-level context features for images or object proposals. 
Besides, the domain query provides a global link between any other local feature tokens, through which any two tokens can attend to each other.
The two global-level feature tokens generated by the two domain queries are further classified by two domain discriminators, respectively. 
The domain discriminators' optimization goal is to make the extracted features more domain-specific and distinguishable, which is opposite to the purpose of the cross-domain model, which tries to learn more domain-invariant features.
Therefore, the domain adversarial training method~\cite{ganin2016domain} is adopted to insert gradient reversal layers and reverse back-propagated gradients from domain discriminators to optimize the detection model to extract domain-invariant features.


\subsubsection{Instance-level Bi-level Graph-based Prototype Alignment (BGPA).}
\begin{figure*}[t]
\begin{center}
	\includegraphics[width=\textwidth]{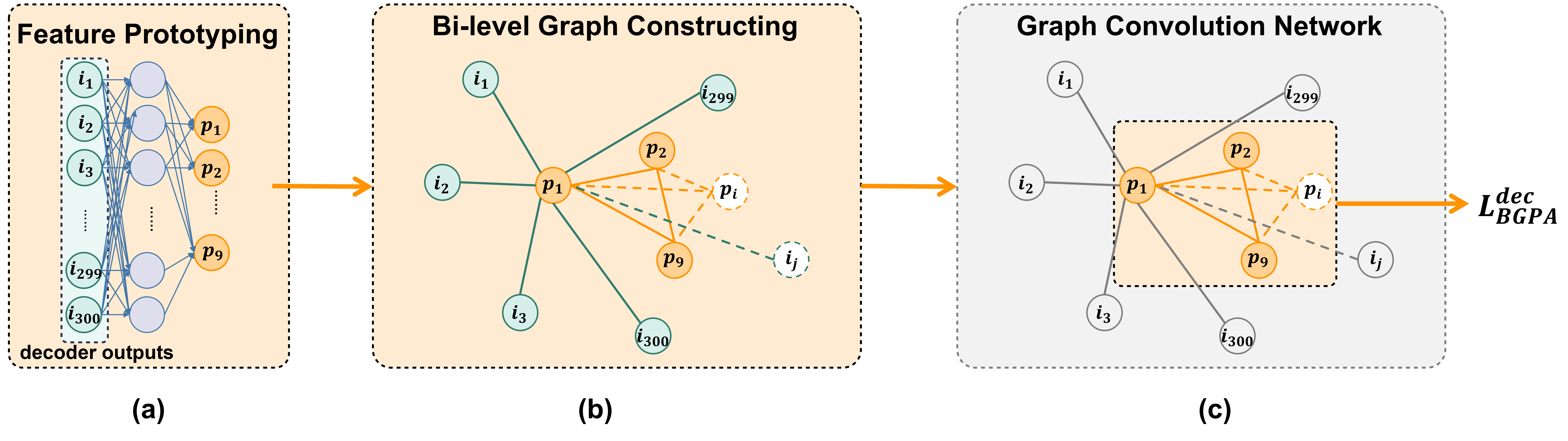}
\end{center}
\vspace{-10pt}
\caption{The proposed bi-level graph-based prototype alignment (BGPA) for instance-level object proposal features. $M$ prototypes (set as 9 in our work) for instance features are first generated with decoder output by an MLP. The prototype features are then aggregated before alignment by constructing a bi-level graph and a GCN.}
\label{fig:BGPA}
\end{figure*}

The above DQFA can effectively shrink global domain gaps in scene layout, but it still suffers from domain shift caused by object visual layout changing. 
Previous work proposes aggregating and further aligning instance features based on their visual similarities~\cite{rezaeianaran2021seeking}. 
In MTTrans, we propose BGPA to build and aggregate prototypes for object features based on their feature similarities with prototyping and graph construction, which can learn from categorical and spatial correlations (since objects in the same category and spatially connected tend to be visually similar~\cite{rezaeianaran2021seeking}) and achieve a more accurate alignment.
As illustrated in Fig.~\ref{fig:BGPA} (a), $M$ prototypes (set as 9 in this work) for object proposals are generated by an MLP with the output feature tokens of the decoder. 
In Fig.~\ref{fig:BGPA} (b), we construct an undirected bi-level graph with the prototypes and the decoder outputs. The prototypes are connected to each other and the decoder outputs, while edges' weights are calculated by the nodes' cosine similarity. 
As shown in Fig.~\ref{fig:BGPA} (c), the created graph is then processed with a graph convolutional network (GCN)~\cite{kipf2016semi,sun2021sugar}. 
Finally, the graph-aggregated prototypes are aligned with a domain discriminator to alleviate domain shift at the instance level.

\subsubsection{Local-level Token-wise Image Feature Alignment (TIFA).}
Though adopting the above two alignments, the mean teacher framework still suffers from the domain shift caused by detailed changes at the local level.
Thus, we additionally add a simple yet effective TIFA that aligns the multi-scale image local feature tokens with domain adversarial training. 
As shown in Fig.~\ref{fig:alignment}, image feature maps of multiple spatial scales are retrieved from different CNN backbone layers and then flattened and concatenated, forming a sequence of local-level image feature tokens.
The feature tokens are aggregated with each other by the cross-scale self-attention mechanism in the encoder, obtaining context-enriched local-level features.
TIFA aligns these tokens one by one with the domain discriminator. 
There exist some more complex feature alignment methods at the local level, and BGPA proposed above can also be applied to these feature tokens. However, these methods may result in performance degradation compared with the simple TIFA, as shown in Section~\ref{sec:4.3}.

\subsection{Progressive Cross-Domain Knowledge Transfer with Mean Teacher and Adversarial Feature Alignment}
\label{sec:3.3}
Compared with previous mean teacher Transformer works that focus on semi-supervised learning, we propose not only the comprehensive multi-level feature alignment described above to strengthen the reliability of the cross-domain pseudo labels but also a two-step progressive transfer training policy. 

First of all, the model needs to learn from the labeled source domain data first and then be able to generate informative pseudo labels. Moreover, it is difficult to simultaneously train a randomly initialized model to perform well on both two distinctly different datasets.
Therefore, we divide the mean teacher training procedure into a burn-in and a transfer training step.
In the first burn-in step, we train the student model in the labeled source domain with object detection and feature alignment tasks. 
The loss function is a combination of both tasks:


\begin{equation}
   \mathcal{L}_{sup} = \mathcal{L}_{det}(I_{src}, y_{src}) -  \mathcal{L}_{adv}
\label{equ:sup_loss}
\end{equation}

Given a source domain image $I_{src} \in \mathcal{D}_{src}$, the backbone $G$ and the encoder $Enc$ of the Transformer produce the image features $f_{src}$, then the decoder $Dec$ produces the features for the object proposals $g$, and finally, the detection head $H$ obtains the bounding boxes and the object category predictions.
The supervised detection loss is defined as follows:

\begin{equation}
    \mathcal{L}_{det}(I_{src}, y_{src}) = l(H(g), y_{src})
\end{equation}


where $l(\cdot)$ denotes the supervised object detection loss which keeps the same with Deformable DETR~\cite{zhu2020deformable}.

The cross-domain adversarial feature alignment loss $\mathcal{L}_{adv}$ comprises four parts: DQFA loss on image features $\mathcal{L}_{enc}^{DQFA}$, DQFA loss on object features $\mathcal{L}_{dec}^{DQFA}$, BGPA loss $\mathcal{L}_{dec}^{BGPA}$, and TIFA loss $\mathcal{L}_{enc}^{TIFA}$:



\begin{equation}
    \mathcal{L}_{adv} = \lambda_1 \mathcal{L}_{enc}^{DQFA}(f)+\lambda_2 \mathcal{L}_{dec}^{DQFA}(g) + \lambda_3 \mathcal{L}_{dec}^{BGPA}(g) + \lambda_4 \mathcal{L}_{enc}^{TIFA}(f)
\label{equ:adv_loss}
\end{equation}

$\mathcal{L}_{enc}^{DQFA}$ is computed in the following way, and the other three parts are computed in the same way:

\begin{equation}
\label{eq:global_adv_loss}
\mathcal{L}_{enc}^{DQFA}(f) = \mathbb{E}_{f \in \mathcal{D}_{src}} \log{D}(f) + \mathbb{E}_{f \in \mathcal{D}_{tgt}} \log(1-{D}(f))
\end{equation}

where $D$ denotes the domain discriminator.

Then, in the transfer training step, both the student and the teacher model are initialized with the model trained in the first step. 
The student model is trained on the source and the target domain alternatively, one epoch for each domain, to maintain the object detection performance and learn to transfer between domains.
The student model is updated with back-propagation, and the teacher model is updated by the student model's weights after each source domain trained epoch. 
The loss for target domain training is made up of two components: UDA object detection loss and adversarial feature alignment loss:
\begin{equation} 
    \mathcal{L}_{unsup} = \mathcal{L}_{det}(I_{tgt}, \hat{y}_{tgt}) -  \mathcal{L}_{adv}
\label{equ:unsup_loss}
\end{equation}
where $\mathcal{L}_{adv}$ is denoted in Eq.~\ref{equ:adv_loss}, $I_{tgt}$ is the unlabeled target domain images, $\hat{y}_{tgt}$ is the generated pseudo labels, and $\mathcal{L}_{det}(I_{tgt}, \hat{y}_{tgt})$ is the pseudo-supervised object detection loss on the target domain.

To summarize, let $T$ denote the object detector and $D$ denote the domain discriminators, the final training objective of MTTrans is defined as :

\begin{equation}
    \min_T \max_D \mathcal{L}_{det}^{src}(T) +  \mathcal{L}_{det}^{tgt}(T)  - \mathcal{L}_{adv}(T,D)
\end{equation}

\section{Evaluation}
In this section, we conduct extensive experiments to demonstrate the advantages of our proposed method. 
In Section~\ref{sec:4.1}, the details of the experimental setup are given. 
In Section~\ref{sec:4.2}, we demonstrate the performance of MTTrans in three challenging domain adaptation scenarios, including Weather, Scene, and Synthetic to Real Adaptation. 
We also conduct comprehensive ablation studies to investigate the impact of each component in Section~\ref{sec:4.3}. 
Finally, we conduct qualitative analysis to provide a better understanding in Section~\ref{sec:4.4}.

\subsection{Experimental Setup}
\label{sec:4.1}
\subsubsection{Datasets}
We evaluate our method on four public datasets, including Cityscapes~\cite{cordts2016cityscapes}, Foggy Cityscapes~\cite{sakaridis2018semantic}, Sim10k~\cite{johnson2016driving}, and BDD100k~\cite{yu2018bdd100k}. We present the performance of MTTrans in three domain adaptation scenarios according to these datasets:

\begin{itemize}
 \item \textbf{Weather adaptation.} In this scenario, we use Cityscapes collected from urban scenes as the source dataset. It consists of 3,475 images with pixel-level annotation, and 2,975 of them are used for training, the other 500 are for evaluation. 
 Foggy Cityscapes constructed from Cityscapes by a fog synthesis algorithm is used as the target dataset. 
 
 \item\textbf{Scene Adaptation.} In this condition, Cityscapes still serves as the source dataset. We utilize the daytime subset of BDD100k as the target dataset, which consists of 36,728 training images and 5,258 validation images annotated with bounding boxes. 
 
 \item \textbf{Synthetic to Real Adaptation.} In this scenario, we utilize Sim10k created by the Grand Theft Auto game engine as the source domain. It is designed to conclude 10,000 training images with 58,701 bounding box annotations for cars. Car instances in Cityscapes are for target domain training and evaluation. 
\end{itemize}

\subsubsection{Implementation Details} 
Our method is built on the basis of Deformable DETR~\cite{zhu2020deformable}. We set ImageNet~\cite{deng2009imagenet} pre-trained ResNet-50~\cite{he2016deep} as CNN backbone in all experiments. In the burn-in step, we adopt Adam optimizer~\cite{kingma2014adam} during training for 50 epochs. We initialize the learning rate as $2 \times 10 ^{-4} $ which decayed by 0.1 after 40 epochs. The batch size is set to 4 for all domain adaptation scenarios. In the second cross-domain training step, the model is trained for 40 epochs, and the initial learning rate is $2 \times 10 ^{-6}$. The learning rate is decayed by 0.1 after 20 epochs. The batch size in this step is set to 2. In addition, we adopt Mean Average Precision (mAP) with a threshold of 0.5 as the evaluation metric. The filtering threshold for pseudo label generation is set to 0.5. All experiments are conducted on NVIDIA Tesla A100 GPUs.

\subsection{Comparisons with Other Methods}
\label{sec:4.2}
The proposed MTTrans framework is assessed in three different domain adaptation scenarios, and the results obtained are compared with prior works in this section. 
SOTA results are achieved in all three scenarios.

\subsubsection{Weather Adaptation}
Variations in weather conditions are common, yet challenging, and object detectors must be reliable under all conditions. As a result, we evaluate the robustness of models under weather changes by transferring from Cityscapes to Foggy Cityscapes. As shown in Table~\ref{tab:1}, MTTrans outperforms other end-to-end methods by a significant margin (43.4\% vs. 41.3\% by the closest and SOTA end-to-end model, SFA). 
In addition, it achieves competitive results when compared with two-stage approaches. 

\begin{table*}[t]
\scriptsize
\begin{center}
\caption{Results of different methods for weather adaptation, that is, from cityscapes to fog cityscapes. FRCNN and DefDETR are abbreviations for Faster R-CNN and Deformable DETR, respectively.}
\begin{tabular}{c|c|cccccccc|c}
\hline
Method                 & Detector & person & rider & car  & truck & bus  & train & mcycle & bicycle & mAP  \\\hline\hline
FasterRCNN~\cite{ren2015faster}(Source)     & FRCNN    & 26.9   & 38.2  & 35.6 & 18.3  & 32.4 & 9.6   & 25.8   & 28.6    & 26.9 \\
DivMatch~\cite{kim2019diversify}      & FRCNN    & 31.8   & 40.5  & 51.0   & 20.9  & 41.8 & 34.3  & 26.6   & 32.4    & 34.9 \\
SWDA~\cite{saito2019strong}           & FRCNN    & 31.8   & 44.3  & 48.9 & 21.0    & 43.8 & 28    & 28.9   & 35.8    & 35.3 \\
SCDA~\cite{zhu2019adapting}          & FRCNN    & 33.8   & 42.1  & 52.1 & 26.8  & 42.5 & 26.5  & 29.2   & 34.5    & 35.9 \\
MTOR~\cite{cai2019exploring}       & FRCNN    & 30.6   & 41.4  & 44.0   & 21.9  & 38.6 & 40.6  & 28.3   & 35.6    & 35.1 \\
CR-DA~\cite{xu2020exploring}         & FRCNN    & 30.0     & 41.2  & 46.1 & 22.5  & 43.2 & 27.9  & 27.8   & 34.7    & 34.2 \\
CR-SW~\cite{xu2020exploring}       & FRCNN    & 34.1   & 44.3  & 53.5 & 24.4  & 44.8 & 38.1  & 26.8   & 34.9    & 37.6 \\
GPA~\cite{xu2020cross}          & FRCNN    & 32.9   & 46.7  & 54.1 & 24.7  & 45.7 & 41.1  & 32.4   & 38.7    & 39.5 \\
ViSGA~\cite{rezaeianaran2021seeking}  & FRCNN    & 38.8   & 45.9  & 57.2 & \textbf{29.9}  & \textbf{50.2} & \textbf{51.9}  & 31.9   & 40.9    & 43.3 \\\hline
FCOS~\cite{tian2019fcos} (Source)           & FCOS     & 36.9   & 36.3  & 44.1 & 18.6  & 29.3 & 8.4   & 20.3   & 31.9    & 28.2 \\
EPM\cite{hsu2020every}         & FCOS     & 44.2   & 46.6  & 58.5 & 24.8  & 45.2 & 29.1  & 28.6   & 34.6    & 39.0   \\\hline
Def DETR~\cite{zhu2020deformable} (Source) & DefDETR  & 37.7   & 39.1  & 44.2 & 17.2  & 26.8 & 5.8   & 21.6   & 35.5    & 28.5 \\
SFA~\cite{wang2021exploring}        & DefDETR  & 46.5   & 48.6  & 62.6 & 25.1  & 46.2 & 29.4  & 28.3   & 44.0      & 41.3\\
MTTrans(Ours)              & DefDETR  & \textbf{47.7}   & \textbf{49.9}  & \textbf{65.2} & 25.8  & 45.9 & 33.8  &  \textbf{32.6}  & \textbf{46.5}      & \textbf{43.4}\\
\hline

\end{tabular}
\label{tab:1}
\end{center}
\vspace{-10pt}
\end{table*}

\begin{table*}[t]
\scriptsize
\begin{center}
\caption{Results of different methods for the scene adaptation, i.e., Cityscapes to BDD100k daytime subset.}
\begin{tabular}{c|c|ccccccc|c}
\hline
Methods                & Detector & person & rider & car  & truck & bus  & mcycle & bicycle & mAP  \\\hline\hline
FasterR-CNN~\cite{ren2015faster}(Source)    & FRCNN    & 28.8   & 25.4  & 44.1 & 17.9  & 16.1 & 13.9   & 22.4    & 24.1 \\
DAF~\cite{chen2018domain}           & FRCNN    & 28.9   & 27.4  & 44.2 & 19.1  & 18.0   & 14.2   & 22.4    & 24.9 \\
SWDA~\cite{saito2019strong}        & FRCNN    & 29.5   & 29.9  & 44.8 & 20.2  & 20.7 & 15.2   & 23.1    & 26.2 \\
SCDA~\cite{zhu2019adapting}         & FRCNN    & 29.3   & 29.2  & 44.4 & 20.3  & 19.6 & 14.8   & 23.2    & 25.8 \\
CR-DA~\cite{xu2020exploring}     & FRCNN    & 30.8   & 29.0    & 44.8 & 20.5  & 19.8 & 14.1   & 22.8    & 26.0   \\
CR-SW~\cite{xu2020exploring}   & FRCNN    & 32.8   & 29.3  & 45.8 & 22.7  & 20.6 & 14.9   & \textbf{25.5}    & 27.4 \\\hline
FCOS~\cite{tian2019fcos}(Source)   & FCOS     & 38.6   & 24.8  & 54.5 & 17.2  & 16.3 & 15.0     & 18.3    & 26.4 \\
EPM~\cite{hsu2020every}      & FCOS     & 39.6   & 26.8  & 55.8 & 18.8  & 19.1 & 14.5   & 20.1    & 27.8 \\\hline
Def DETR~\cite{zhu2020deformable}(Source) & DefDETR  & 38.9   & 26.7  & 55.2 & 15.7  & 19.7 & 10.8   & 16.2    & 26.2 \\
SFA~\cite{wang2021exploring}  & DefDETR  & 40.2   & 27.6  & 57.5 & 19.1  & 23.4 & 15.4   & 19.2    & 28.9\\
MTTrans(Ours)  & DefDETR  & \textbf{44.1}   & \textbf{30.1}  & \textbf{61.5} & \textbf{25.1}  & \textbf{26.9} & \textbf{17.7}   & 23.0    & \textbf{32.6}\\\hline
\end{tabular}
\label{tab:3}
\end{center}
\vspace{-15pt}
\end{table*}

\subsubsection{Scene Adaptation}
Scene layouts are not static and frequently change in real-world applications, especially in autonomous driving scenarios. As a result, model performance under scene adaptation is critical. As shown in Table~\ref{tab:3}, MTTrans obtains SOTA results (32.6\%) with significant improvements over previous works. In addition, the performance of six out of seven categories in the target domain dataset has been improved.
(As the SOTA method in the weather adaptation scenario ViSGA~\cite{rezaeianaran2021seeking} did not open-source their code and did not report performance in the scene adaptation scenario, it is not included here.)

\subsubsection{Synthetic to Real Adaptation} 
Images and their corresponding annotations can be created by video games such as GTA, which can considerably minimize the manual cost of data collection and annotating. As a result, it is worthwhile to enable the object detector to learn from synthetic images and then adapt to generic real-world images. Therefore, as shown in Table~\ref{tab:2}, we test MTTrans in the synthetic to real adaptation scenario, obtaining an accuracy of 57.9\%, outperforming the previous state-of-the-art by 5.3\%.

\begin{table} [t]
\scriptsize
\begin{center}
\caption{Results of different methods for the synthetic to real
adaptation, i.e. Sim10k to Cityscapes.}
\begin{tabular}{c|c|c}
\hline
Methods                & Detector & carAP \\\hline\hline
FasterRCNN~\cite{ren2015faster}(Source)     & FRCNN    & 34.6  \\
DAF~\cite{chen2018domain}    & FRCNN    & 41.9  \\
DivMatch~\cite{kim2019diversify}      & FRCNN    & 43.9  \\
SWDA~\cite{saito2019strong}          & FRCNN    & 44.6  \\
SCDA~\cite{zhu2019adapting}         & FRCNN    & 45.1  \\
MTOR~\cite{cai2019exploring}           & FRCNN    & 46.6  \\
CR-DA~\cite{xu2020exploring}         & FRCNN    & 43.1  \\
CR-SW~\cite{xu2020exploring}          & FRCNN    & 46.2  \\
GPA~\cite{xu2020cross}            & FRCNN    & 47.6  \\
ViSGA~\cite{rezaeianaran2021seeking}       & FRCNN    & 49.3  \\\hline
FCOS~\cite{tian2019fcos}(Source)           & FCOS     & 42.5  \\
EPM~\cite{hsu2020every}           & FCOS     & 47.3  \\\hline
DefDETR~\cite{zhu2020deformable}(Source) & DefDETR  & 47.4  \\
SFA~\cite{wang2021exploring} & DefDETR  & 52.6\\
MTTrans(Ours)              & DefDETR  & \textbf{57.9} \\\hline
\end{tabular}
\label{tab:2}
\end{center}
\vspace{-15pt}
\end{table}

\subsection{Ablation Studies}
\label{sec:4.3}

\begin{table} [t]
\centering
\scriptsize
\caption{Ablation studies on the weather adaptation scenario, with Cityscapes to Foggy Cityscapes. MT stands for the mean teacher framework and SharedQE denotes the shared object queries of decoder inputs. 
Components of other models that differ from MTTrans are marked in red.}
\begin{tabular}{cccccccccc}
\hline
 &
   &
   &
  \multicolumn{2}{c}{DQFA} &
  \multicolumn{2}{c}{TIFA} &
  \multicolumn{2}{c}{BGPA} &
   \\ \cline{4-9}
\multirow{-2}{*}{Methods} &
  \multirow{-2}{*}{MT} &
  \multirow{-2}{*}{SharedQE} &
  enc &
  dec &
  enc &
  dec &
  enc &
  dec &
  \multirow{-2}{*}{mAP50} \\ \hline
Deformable DETR (Source) &
  {\color[HTML]{FF0000} \XSolidBrush} &
  {\color[HTML]{FF0000} \XSolidBrush} &
  {\color[HTML]{FF0000} \XSolidBrush} &
  {\color[HTML]{FF0000} \XSolidBrush} &
  {\color[HTML]{FF0000} \XSolidBrush} &
  \XSolidBrush &
  \XSolidBrush &
  {\color[HTML]{FF0000} \XSolidBrush} &
  28.5 \\
MTTrans-AS0(MT-DefDETR) &
  \Checkmark &
  \Checkmark &
  {\color[HTML]{FF0000} \XSolidBrush} &
  {\color[HTML]{FF0000} \XSolidBrush} &
  {\color[HTML]{FF0000} \XSolidBrush} &
  \XSolidBrush &
  {\color[HTML]{333333} \XSolidBrush} &
  {\color[HTML]{FF0000} \XSolidBrush} &
  35.843 \\ \hline
MTTrans-AS11 &
  \Checkmark &
  \Checkmark &
  {\color[HTML]{FF0000} \XSolidBrush} &
  \Checkmark &
  \Checkmark &
  {\color[HTML]{333333} \XSolidBrush} &
  {\color[HTML]{333333} \XSolidBrush} &
  \Checkmark &
  43.183 \\
MTTrans-AS12 &
  \Checkmark &
  \Checkmark &
  \Checkmark &
  {\color[HTML]{FF0000} \XSolidBrush} &
  \Checkmark &
  {\color[HTML]{333333} \XSolidBrush} &
  {\color[HTML]{333333} \XSolidBrush} &
  \Checkmark &
  42.962 \\
MTTrans-AS13 &
  \Checkmark &
  \Checkmark &
  \Checkmark &
  \Checkmark &
  {\color[HTML]{FF0000} \XSolidBrush} &
  {\color[HTML]{333333} \XSolidBrush} &
  {\color[HTML]{333333} \XSolidBrush} &
  \Checkmark &
  43.013 \\
MTTrans-AS14 &
  \Checkmark &
  \Checkmark &
  \Checkmark &
  \Checkmark &
  \Checkmark &
  {\color[HTML]{333333} \XSolidBrush} &
  {\color[HTML]{333333} \XSolidBrush} &
  {\color[HTML]{FF0000} \XSolidBrush} &
  43.003 \\
MTTrans-AS15 &
  \Checkmark &
  {\color[HTML]{FF0000} \XSolidBrush} &
  \Checkmark &
  \Checkmark &
  \Checkmark &
  {\color[HTML]{333333} \XSolidBrush} &
  {\color[HTML]{333333} \XSolidBrush} &
  \Checkmark &
  42.797 \\
MTTrans-AS21 &
  \Checkmark &
  \Checkmark &
  \Checkmark &
  \Checkmark &
  \Checkmark &
  {\color[HTML]{FF0000} \Checkmark} &
  {\color[HTML]{333333} \XSolidBrush} &
  {\color[HTML]{FF0000} \XSolidBrush} &
  42.953 \\
MTTrans-AS22 &
  \Checkmark &
  \Checkmark &
  \Checkmark &
  \Checkmark &
  {\color[HTML]{FF0000} \XSolidBrush} &
  {\color[HTML]{333333} \XSolidBrush} &
  {\color[HTML]{FF0000} \Checkmark} &
  \Checkmark &
  43.068 \\ \hline
MTTrans &
  \Checkmark &
  \Checkmark &
  \Checkmark &
  \Checkmark &
  \Checkmark &
  \XSolidBrush &
  \XSolidBrush &
  \Checkmark &
  43.413 \\ \hline
\end{tabular}
\label{tab:4}
\vspace{-15pt}
\end{table}

To better analyze each component in our proposed MTTrans framework, we conduct ablation studies by removing parts of the components in MTTrans. From Table~\ref{tab:4}, we can observe: (1) Adding the mean teacher framework directly to Deformable DETR (MTTrans-AS0, MT-DefDETR)) can improve its performance on the target domain (+7.34 mAP). We can notice that it is critical to introduce the mean teacher framework in cross-domain adaptation, but there is still much space for improvement due to the poor quality of pseudo labels. (2) Removing any aspect of MTTrans (MTTrans-AS11 to MTTrans-AS14)will result in performance degradation. Removing DQFA for the encoder (MTTrans-AS11) results in the slightest performance drop (-0.230mAP), while removing DQFA for the decoder (MTTrans-AS12) results in the highest drop (-0.451 mAP); (3) Altering the alignment technique for the decoder from BGPA to TIFA (MTTrans-AS21, -0.460 mAP), or replacing TIFA for the encoder with BGPA (MTTrans-AS22, -0.345 mAP) both result in performance drop; (4) Removing the shared object queries between the teacher and student models (MTTrans-AS15) will also decrease MTTrans's performance (-0.616 mAP). 

\subsection{Visualization and Analysis}
\label{sec:4.4}

\subsubsection{Pseudo Label Visualization}

\begin{figure*}[t]
\begin{center}
	\includegraphics[width=\textwidth]{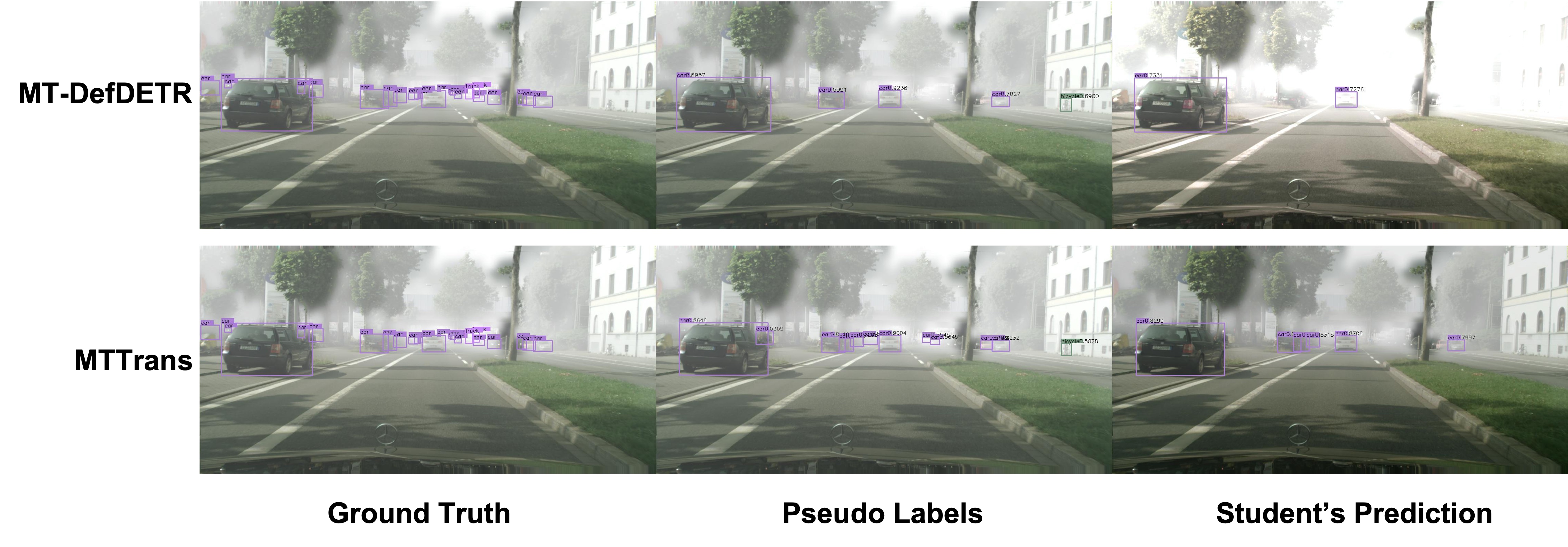}
\end{center}
\caption{Visualization results of the generated pseudo labels, ground truth annotations, and student model predictions. As seen in the visualization result, MTTrans can generate pseudo labels of higher quality compared with MT-DefDETR; and the teacher model performs better than the student model. MT-DefDETR stands for directly applying the mean teacher framework to Deformable DETR.}
\label{fig:pseudo_label_vis}
\end{figure*}

We show some visualization results of the generated pseudo labels and student model's prediction, produced by MTTrans and the mean teacher version of Deformable DETR, as shown in Fig.~\ref{fig:pseudo_label_vis}. It should be noted that input images are augmented randomly during training, and the student model's performance can not be compared between the two model's predictions in the figure. However, the teacher model's inputs are almost the same, and the pseudo labels generated by MT-DefDETR are of lower quality than MTTrans's. Furthermore, the pseudo labels generated by the teacher model are better than the student model's predictions, ensuring that the student model can learn to improve via pseudo labels.

\begin{figure} [t]
	\centering
	\includegraphics[width=\linewidth]{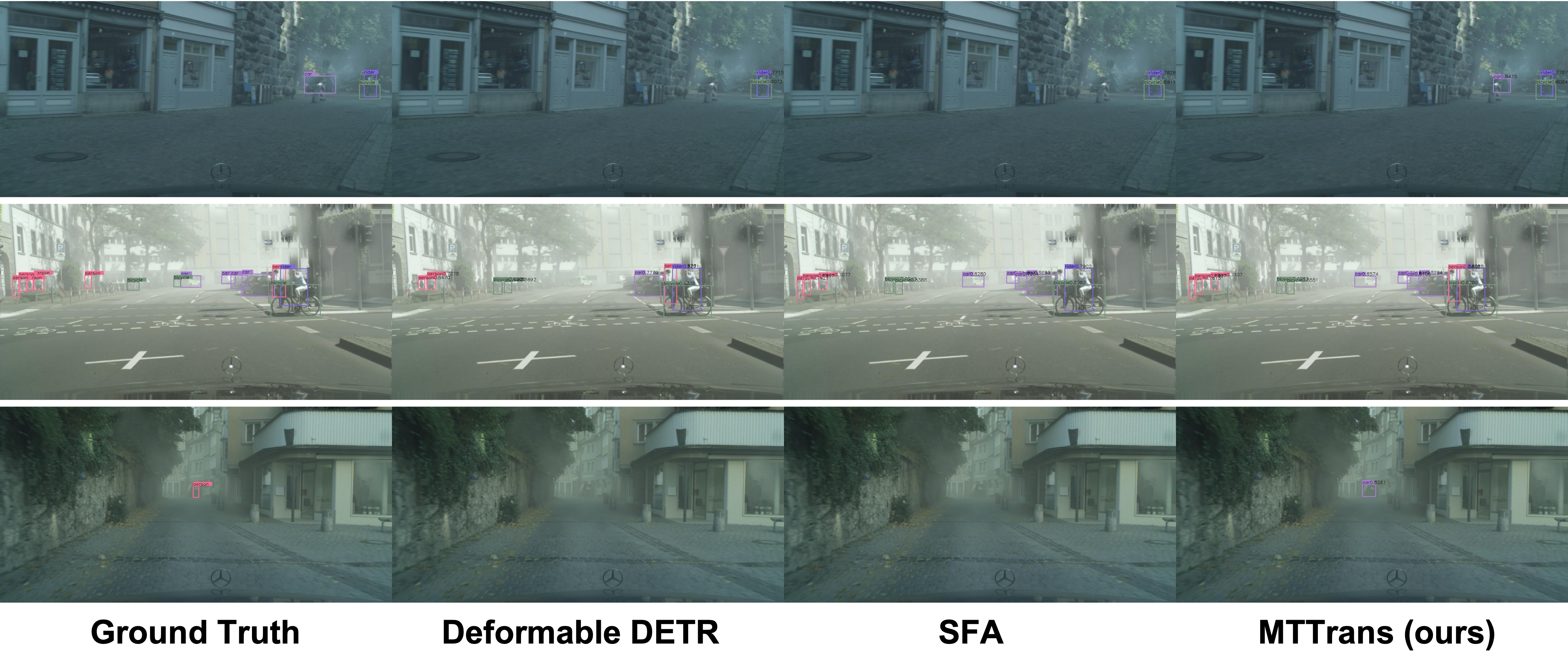}
	\caption{Visualization of detection results on the Foggy Cityscapes dataset, from left to right, are ground truth, results obtained by Deformable DETR, SFA, and MTTrans. The predicted category and prediction confidence can be seen in the bounding box labels. Recommend to read with computers, and the original image files are attached with supplement materials.}
	\label{fig:foggy_vis}
	\centering
\end{figure}



\subsubsection{Detection Results}
We show some visualization results of MTTrans on the Foggy Cityscapes dataset, accompanied by ground truth, baseline, and previous state-of-the-art (SOTA) methods. As shown in Fig.~\ref{fig:foggy_vis}, MTTrans can better detect complex objects covered by fog in the distance compared with Deformable DETR and SFA. The results are consistent with the numerical assessment results, indicating that MTTrans manages to mitigate the domain shift problem in the UDA Transformer detector. 
Furthermore, as detection confidence scores are shown in the left-top corner of the bounding box (best zoom in $\times 8$), MTTrans can slightly boost the scores of true positive predictions, further indicating its strength.
As shown in the third row of Fig.\ref{fig:foggy_vis}, MTTrans manages to retrieve objects ignored by SFA but discovered by Deformable DETR. From the last row in Fig.~\ref{fig:foggy_vis}, even though our proposed MTTrans fails to discover the blurry and indistinguishable person, it succeeds in detecting cars that are not labeled in the ground truth. More visualization of other datasets is presented in the supplement materials.

\section{Conclusions}
This paper introduces an end-to-end cross-domain object detection Transformer named MTTrans based on the mean teacher framework and the comprehensive multi-level feature alignment.
The mean teacher framework is adopted in MTTrans to fully utilize the unlabeled target domain dataset, making it pseudo-labeled and available in the training of the object detection task.
To compensate for the lack of explicit cross-domain ability of the mean teacher framework and to further improve the quality of the generated pseudo labels under domain shift, we propose the multi-level feature alignment which aligns image and object features at local, instance, and global levels based on the cross-scale self-attention mechanism in Deformable DETR~\cite{zhu2020deformable}, including domain query-based feature alignment for global-level image and object features, bi-level graph-based prototype alignment for instance-level object features, and token-wise image feature alignment for local-level image features, respectively.
Experimental results demonstrate the effectiveness of our MTTrans. We hope that this paper can inspire future work on cross-domain object detection.

\subsubsection{Acknowledgment.}
We thank Xiaoqi Li and Zhaoqing Wang for their help with the manuscript revision and the anonymous reviewers for their helpful comments to improve the paper. The authors of this paper are supported by the NSFC through grant No.U20B2053, and we also thanks the support from Beijing Advanced Innovation Center for Future Blockchain and Privacy Computing.

\clearpage
%
%
\bibliographystyle{splncs04}
\bibliography{egbib}
\end{document}